\documentclass[conference, a4paper]{IEEEtran}
\IEEEoverridecommandlockouts
\usepackage{amsmath,amssymb,amsfonts}
\usepackage{graphicx}
\usepackage{textcomp}
\usepackage{xcolor}
\usepackage{amsthm}
\usepackage{amssymb}
\usepackage{setspace}
\usepackage{hyperref}
\usepackage{psfrag}
\usepackage{booktabs}
\usepackage{adjustbox}
\usepackage{multirow}
\newtheorem{theorem}{Theorem}

\newcommand{\bd}[1]{\mathbf{#1}}
\newcommand{\set}[1]{\mathcal{#1}}

\usepackage[english]{babel}

\usepackage{algorithm}
\usepackage{algpseudocode}

\algrenewcommand\algorithmicrequire{\textbf{Input:}}
\algrenewcommand\algorithmicensure{\textbf{Output:}}
\algrenewcommand\algorithmicforall{\textbf{for}}

\usepackage{comment}
\usepackage{subcaption}
\usepackage{float}

\begin{document}
\title{Multi-task Online Learning for Probabilistic\\ Load Forecasting}

 \author{\IEEEauthorblockN{Onintze Zaballa\IEEEauthorrefmark{1}, Ver\'onica \'Alvarez\IEEEauthorrefmark{1}, Santiago Mazuelas\IEEEauthorrefmark{1}\IEEEauthorrefmark{2}} \IEEEauthorblockA{\IEEEauthorrefmark{1}Basque Center for Applied Mathematics (BCAM)}\IEEEauthorblockA{\IEEEauthorrefmark{2}IKERBASQUE - Basque Foundation for Science\\ Email: \{ozaballa, valvarez, smazuelas\}@bcamath.org  }
 \thanks{Funding in direct support of this work has been provided by projects  ID2022-137063NB-I00,  CNS2022-135203, and CEX2021-001142-S funded by MCIN/AEI/10.13039/501100011033 and the European Union “NextGenerationEU”/PRTR, and programes ELKARTEK and BERC-2022-2025 funded by the Basque Government.}}

\maketitle
\begin{abstract}
Load forecasting is essential for the efficient, reliable, and cost-effective management of power systems. Load forecasting performance can be improved by learning the similarities among multiple entities (e.g., regions, buildings). Techniques based on multi-task learning obtain predictions by leveraging consumption patterns from the historical load demand of multiple entities and their relationships. However, existing techniques cannot effectively assess inherent uncertainties in load demand or account for dynamic changes in consumption patterns. This paper proposes a multi-task learning technique for online and probabilistic load forecasting. This technique provides accurate probabilistic predictions for the loads of multiple entities by leveraging their dynamic similarities. The method's performance is evaluated using datasets that register the load demand of multiple entities and contain diverse and dynamic consumption patterns. The experimental results show that the proposed method can significantly enhance the effectiveness of current multi-task learning approaches across a wide variety of load consumption scenarios.
\end{abstract}


\section{Introduction}

Load forecasting is essential for the efficient, reliable, and cost-effective management of power systems \cite{wang2023probabilistic}. Accurate load forecasting can significantly improve energy management, as electrical systems are required to maintain a real-time balance between power generation and demand.~\cite{wang2023multitask}. Predicting load demand using disaggregated data, such as individual households or residential areas, is generating increasing interest with the advent of smart grids and meters \cite{wang2018review}. However, load forecasting remains a challenging problem in such scenarios where energy demand can heavily fluctuate due to uncertain factors such as evolving consumer behaviors \cite{li2023residential}.

\begin{figure}
    \centering
        \psfrag{Single}[t][t][1]{Single-task learning}
        \psfrag{Multi}[t][t][1]{Multi-task learning}
        \psfrag{A}[][l][0.8]{ \hspace{1cm} Entity 1}
        \psfrag{B}[][l][0.8]{ \hspace{1cm}  Entity 2}
        \psfrag{C}[][l][0.8]{ \hspace{1cm} Entity 3}
        \includegraphics[width=0.45\textwidth]{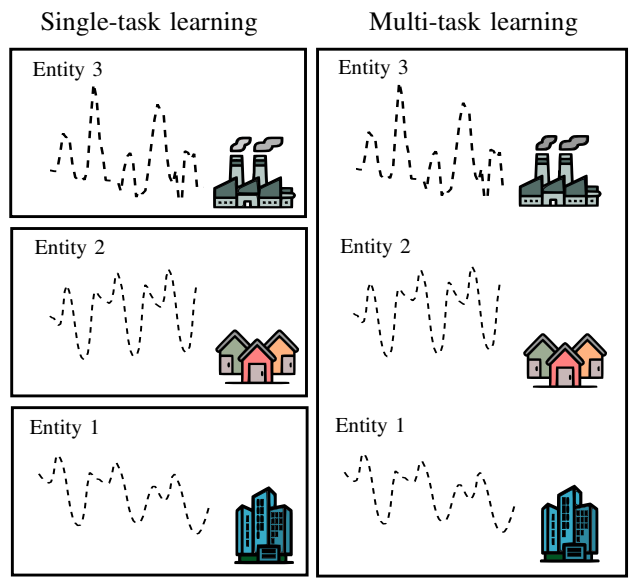}
    \caption{Consumption patterns of different entities can be similar (entity 1 and entity 2). Single-task learning techniques can only leverage information from the corresponding entity, while multi-task learning techniques use information from multiple entities.}
    \label{fig:intro}
\end{figure}

The load forecasting performance can be improved by leveraging the similarities among multiple entities (e.g., regions, buildings) (see Figure~\ref{fig:intro}) \cite{kim2024spatial}. Forecasting techniques based on multi-task learning enable to jointly learn from multiple related tasks~\cite{zhang2021survey}. In load forecasting, tasks can be referred to as entities, such as regions, buildings, or apartments. Techniques based on multi-task Gaussian processes have been developed to learn correlations based on load consumption patterns \cite{Zhang2014}. More recently, a multi-task learning algorithm for a Bayesian spatiotemporal Gaussian process model has been proposed to assess the relationship among various residential areas based on the effects of time-varying environmental and traffic conditions on load consumption \cite{gilanifar2019multitask}. In addition, kernel-based multi-task learning techniques have been used for middle-term load forecasting by leveraging correlations in seasonal patterns \cite{fiot2016electricity}. Such techniques have demonstrated improved accuracy compared to single-task learning techniques by considering that the consumption patterns of one entity may affect the predictions of another entity. 

Most conventional techniques for single-task and multi-task load forecasting provide single-value predictions based on offline learning \cite{amjady2001short, espinoza2007electric, fiot2016electricity}. These methods face challenges due to the inherent uncertainty in load demand and time-varying changes in consumption patterns. Probabilistic forecasting techniques can provide information about the probability of different outcomes and achieve more accurate forecasts in uncertain scenarios. Existing probabilistic techniques (e.g., quantile regression \cite{liu2015probabilistic}) generate forecasts and assess load uncertainties, providing a measure of confidence in the predictions. Online learning techniques can dynamically adapt to changes in load consumption patterns. Existing online learning methods adapt to dynamic changes in load consumption patterns by regularly retraining models of conventional techniques~\cite{paarmann1995adaptive}, adjusting predictions with new observations \cite{obst2021adaptive}, or updating weights of combined forecasting techniques~\cite{von2020online}. In the literature, online learning techniques provide single-value load predictions \cite{laouafi2017online, ba2012adaptive}, while probabilistic load forecasting methods are based on offline learning~\cite{liu2015probabilistic, yang2019bayesian}.

This paper extends the probabilistic and online learning methodology in \cite{alvarez2021probabilistic} for multi-task load forecasting. The proposed method provides accurate probabilistic predictions for loads of multiple entities based on  their time-varying relationship in consumption patterns. The method is designed to dynamically adapt to changing load patterns and dependencies among entities over time. The main contributions of this paper are as follows.

\begin{itemize}\itemsep0em 
    \item We develop online learning techniques for multi-task load forecasting based on vector-valued hidden Markov models (HMMs).
    \item We develop forecasting techniques that obtain probabilistic predictions for multiple tasks using the most recent HMM parameters.
    \item We assess the performance of the proposed method compared to existing multi-task techniques for load forecasting.
\end{itemize}

The rest of this paper is organized as follows. Section~\ref{sec:loadforecasting} describes the load forecasting problem and introduces the preliminaries for understanding the proposed method. In Section \ref{sec:multitask}, we outline the proposed method's learning and prediction techniques. Section \ref{sec:experiments} evaluates the performance of the proposed method compared to existing techniques. Finally, Section \ref{sec:conclusion} draws the conclusions.

\section{Load forecasting}
\label{sec:loadforecasting}
This section first describes the problem formulation. Then we provide an overview of the preliminaries necessary for developing the multi-task learning method.

\subsection{Problem formulation}
Load forecasting methods aim to estimate future loads using historical consumption data and observations that influence future loads, such as the hour of the day, the day of the week, or weather forecasts. Load forecasting techniques learn a predictive function that relates instance vectors $\bd{x}$ to target vectors $\bd{y}$. 

Instance vectors $\bd{x}$ (predictors) are composed of past loads and observations related to future loads, and target vectors $\bd{y}$ (responses) are composed of future loads. Loads are denoted by $s$ and load forecasts by~$\widehat{s}$. In addition, $s_t\in \mathbb{R}$ denotes the load for one specific entity at time $t$ and $\widehat{s}_t\in \mathbb{R}$ its load forecast. For each time $t$, the observations vector related to one specific entity is denoted by $\bd{r}_t\in \mathbb{R}^N$, which includes data such as weather forecasts. Each time $t$ has an associated calendar variable $c(t) \in \{1,2,...,C\}$ that describes time-related factors that may affect load consumption, such as the hour of the day, day of the week, or holiday period.

Single-task learning techniques predict future loads of one entity (e.g., region, building) based on its past load consumption and observations that influence future loads \cite{alvarez2021probabilistic}. For a prediction horizon $L$ (e.g., 24 hours, 60 minutes) and prediction times $t+1, t+2, ..., t+L$, the instance vector is defined as $\bd{x} = [s_{t-m}, s_{t-m+1} ..., s_t, \bd{r}_{t+1},\bd{r}_{t+2},..., \bd{r}_{t+L} ]^\top$, the target vector is defined as $\bd{y}=[s_{t+1}, s_{t+2}, ..., s_{t+L}]^\top$ and the vector of load forecast is given by $\widehat{\bd{y}}=[\widehat{s}_{t+1},\widehat{s}_{t+2},...,\widehat{s}_{t+L}]$. Conventional techniques such as Linear Regression \cite{espinoza2007electric} use instance vectors composed of historical loads, temperature-related variables and calendar information. Such algorithms obtain load forecasts as $\widehat{\bd{y}}=\bd{W} \bd{x} +\textbf{b}$, where matrix $\bd{W}$ and vector~$\bd{b}$ are learned using a set of training samples $\{(x_i,y_i)\}_{i=1}^{t_0}$ formed by pairs of previous instance-target vectors.

In a multi-task learning framework, the goal is to estimate future loads of $K$ entities simultaneously. These methods account not only for past loads and observations that influence future loads, but also for the shared consumption patterns among the entities. In particular, load consumption patterns of each entity are used to predict future loads of the other entities. The vector of loads of $K$ entities is denoted by $\bd{s}= [s_1,s_2,...,s_K]^\top$ and load forecasts by \mbox{$\widehat{\bd{s}}=[ \widehat{s}_1, \widehat{s}_2,...,\widehat{s}_K]^\top$}, with $\bd{s}_t\in \mathbb{R}^K$ and $\widehat{\bd{s}}_t\in \mathbb{R}^K$ being the loads and the loads forecast at time $t$. In addition, we denote by $\bd{r}_t=[ \bd{r}_{t,1}^\top, \bd{r}_{t,2}^\top,...,\bd{r}_{t,K}^\top]^\top\in \mathbb{R}^{K N}$ the array composed of concatenating observation vectors of the $K$ entities, where $\bd{r}_{t,i}$ denotes the observation vector at time $t$ for entity \mbox{$i\in \{1,2,...,K\}$}. For a prediction horizon $L$ and prediction times $t+1,t+2, ..., t+L$, the instance vector is defined as $\bd{x} = [\bd{s}_{t-m}^\top, \bd{s}_{t-m+1}^\top,..., \bd{s}_t^\top, \bd{r}_{t+1},\bd{r}_{t+2}, ..., \bd{r}_{t+L} ]^\top$, the target vector is defined as $\bd{y} = [\bd{s}_{t+1}^\top,\bd{s}_{t+2}^\top,...,\bd{s}_{t+L}^\top]^\top$, and the vector of load forecast as $\widehat{\bd{y}} = [\widehat{\bd{s}}_{t+1}^\top,\widehat{\bd{s}}_{t+2}^\top,...,\widehat{\bd{s}}_{t+L}^\top]^\top$.

\subsection{Preliminaries}
A single-task online learning technique is proposed in \cite{alvarez2021probabilistic} based on HMMs, which models the relationship between loads and observations for an individual load entity. The following describes the main components of such a method. The details can be found in \cite{alvarez2021probabilistic}.

The HMM is defined by the following two conditional distributions: $p(s_t|s_{t-1})$, which characterizes the relationship between consecutive loads; and $p(\bd{r}_t|s_t)$, which characterizes the relationship between the loads and the observation vector. In particular, the conditional distributions $p(s_t|s_{t-1})$ and $p(\bd{r}_t|s_t)$ are modeled using Gaussian distributions with mean $\bd{u}^\top\boldsymbol{\eta}$ and standard deviation $\sigma$, where $\bd{u}$ is a known feature vector. The parameters $\boldsymbol{\eta}$ and $\sigma$ vary for each calendar type $c(t) ~\in~\{1,2,...,C\}$ and change over time. 

For each calendar type $c=c(t)$, $\bd{u}_s^\top\boldsymbol{\eta}_{s,c}$ denotes the mean and $\sigma_{s,c}$ the standard deviation that determine the conditional distribution of loads at time $t$ given the loads at time $t-1$, that is,

\begin{equation}
  p(s_{t}| s_{t-1}) = \set{N}(s_t;  \bd{u}_{s}^\top \boldsymbol{\eta}_{s,c} , \sigma_{s,c}) 
  \label{eq:model1}
\end{equation}
where $\boldsymbol{\eta}_{s,c}\in \mathbb{R}^{2}$,  $\sigma_{s,c} \in \mathbb{R}$, and $\bd{u}_{s} = [1, s_{t-1} ]^\top$. In addition, for each $c=c(t)$, $\bd{u}_r^\top \boldsymbol{\eta}_{r,c}$ and $\sigma_{r,c}$ denote the mean and standard deviation that characterize the conditional distribution of loads at time $t$ and the observation vectors. Assuming no prior information is available for the loads, we have that
\begin{equation}
    p(\bd{r}_{t}| s_{t}) \propto p(s_{t}| \bd{r}_{t}) = \set{N}(s_t;  \bd{u}_r^\top \boldsymbol{\eta}_{r,c}, \sigma_{r,c})
    \label{eq:model2}
\end{equation}
where $\boldsymbol{\eta}_{r,c}\in \mathbb{R}^{R}$,  $\sigma_{r,c} \in \mathbb{R}$, and $\bd{u}_r = u_{r}(\bd{r}_t) \in \mathbb{R}^{R}$. 


Parameters $\boldsymbol{\eta}$ and $\sigma$ in \eqref{eq:model1} and \eqref{eq:model2} are updated every time new loads and observations corresponding to calendar type $c$ are obtained. Specifically, if $s_{t_1}, s_{t_2}, ..., s_{t_n}$ are loads obtained at times with calendar type $c \in \{1, 2, ..., C\}$, that is, \mbox{$c = c(t_1) = c(t_2) = ... = c(t_n)$}, and $\bd{u}_{t_1}, \bd{u}_{t_2} , ..., \bd{u}_{t_n}$ are their corresponding feature vectors, the following recursions enable the adaptive online learning of the parameters $\boldsymbol{\eta}_{s,c}$, $\boldsymbol{\eta}_{r,c}$, $\sigma_{s,c}$ and $\sigma_{r,c}$, as follows \cite{alvarez2021probabilistic}.
\begin{align}
\label{eq:eta_i}
\boldsymbol{\eta}_i & = \boldsymbol{\eta}_{i-1} + \frac{\left(s_{t_i} - \bd{u}_{t_i}^\top\boldsymbol{\eta}_{i-1} \right) \bd{u}_i^\top \bd{P}_{i-1}}{\lambda + \bd{u}_{t_i}^\top \bd{P}_{i-1} \bd{u}_{t_i}} \\
\label{eq:sigma_i}
\sigma_{i} & = \sqrt{\sigma_{i-1}^2 -  \frac{1}{\gamma_{i}} \left(\sigma_{i-1}^2-\frac{\lambda^2 \left(s_{t_i} - \bd{u}_{t_i}^\top\boldsymbol{\eta}_{i-1}\right)^2 }{(\lambda + \bd{u}_{t_i}^\top \bd{P}_{i-1} \bd{u}_{t_i})^2}\right)}
\end{align}
with 
\begin{align}
\label{eq:P_i}
    \bd{P}_i = & \frac{1}{\lambda} \left( \bd{P}_{i-1} - \frac{\bd{P}_{i-1} \bd{u}_{t_i} \bd{u}_{t_i}^\top \bd{P}_{i-1}}{ \lambda + \bd{u}_{t_i}^\top \bd{P}_{i-1} \bd{u}_{t_i}} \right)\\
\label{eq:gamma_i}
\gamma_{i}  = & \lambda \gamma_{i-1} + 1
\end{align}
where $\lambda \in (0,1]$ is a forgetting factor that increases the influence of the most recent data. The parameters are updated using the values of the previous time, and the state variables are given by~\eqref{eq:P_i} and~\eqref{eq:gamma_i}.

After updating the parameters with such recursions, the method also enables the recursive estimation of probabilistic load forecasts $\set{N}(s_{t+i}; \widehat{s}_{t+i}, \widehat{e}_{t+i})$, for $i=1,2,...,L$. Such forecasts are obtained with the most recent parameters every time new instance vectors $\bd{x}$ arrive. See \cite{alvarez2021probabilistic} for more details.

The results in this section describe the learning and prediction steps for the single-task probabilistic forecasting method based on online learning. The following section extends this methodology to the multi-task learning framework.

\section{Multi-task probabilistic load forecasting based on online learning}
\label{sec:multitask}

This section describes the presented techniques for probabilistic and online multi-task load forecasting. The proposed method is based on vector-valued HMMs that model the relationship between load consumption and observations for multiple entities.

The multi-task HMM is defined by the conditional distributions $p(\bd{s}_t|\bd{s}_{t-1})$, which characterizes the relationship between multiple consecutive loads, and $p(\bd{r}_t|\bd{s}_t)$, which characterizes the relationship between the loads and the observation vectors of the entities, with $\bd{s}_t=[s_{t,1}, s_{t,2},..., s_{t,K}]^\top \in \mathbb{R}^{K}$ and $\bd{r}_t=[ \bd{r}_{t,1}^\top, \bd{r}_{t,2}^\top,...,\bd{r}_{t,K}^\top]^\top \in \mathbb{R}^{KN}$. These conditional distributions are modeled using multivariate Gaussian distributions with mean $\bd{M}\bd{u}$ and covariance matrix $\bd{\Sigma}$, where $\bd{u}$ is a known feature vector. The parameters described by matrices $\bd{M}$ and $\bd{\Sigma}$ vary for each calendar type $c(t) \in \{1,2,...,C\}$ and change over time. The covariance matrix $\bd{\Sigma}$ provides information on how one task can influence and be influenced by another task at each time.

For each calendar type $c=c(t)$, we denote by $\bd{M}_{s,c} \bd{u}_s$ and $\bd{\Sigma}_{s,c}$ the mean and covariance matrix that characterize the conditional distribution of loads at time $t$ given the loads at time $t-1$, that is,
\begin{equation}
\label{eq:model1_MT}
  p(\bd{s}_{t}| \bd{s}_{t-1}) = \set{N}(\bd{s}_t;   \bd{M}_{s,c} \bd{u}_{s}, \bd{\Sigma}_{s,c}). 
  \end{equation}
where $\bd{M}_{s,c}~\in~\mathbb{R}^{K \times (K+1)}$, $\bd{\Sigma}_{s,c} \in \mathbb{R}^{K \times K}$, and $\bd{u}_{s}~=~[1, s_{t-1,1}, s_{t-1,2} , ..., s_{t-1,K} ]^\top$.

For each calendar type $c=c(t)$, we denote by $\bd{M}_{r,c} \bd{u}_r$ and $\bd{\Sigma}_{r,c}$ the mean and covariance matrix that characterize the conditional distribution of loads at time $t$ and the observation vectors. Therefore, assuming no prior information is available for the loads, we have that
\begin{equation}
\label{eq:model2_MT}
    p(\bd{r}_{t}| \bd{s}_{t}) \propto p(\bd{s}_{t}| \bd{r}_{t}) = \set{N}(\bd{s}_t;  \bd{M}_{r,c} \bd{u}_r, \bd{\Sigma}_{r,c})
\end{equation}
where $\bd{M}_{r,c}\in \mathbb{R}^{K \times KR}$,  $\bd{\Sigma}_{r,c} \in \mathbb{R}^{K \times K}$, and $\bd{u}_r~=~u_{r}(\bd{r}_t)~\in \mathbb{R}^{KR}$, which is the function that represents the observations.

\subsection{Learning}
In the following, we extend the results of single-task learning to handle multiple entities simultaneously. The multi-task HMM for each time $t$ is characterized by parameters 
\begin{equation}
    \Theta = \{ \bd{M}_{s,c}, \boldsymbol{\Sigma}_{s,c}, \bd{M}_{r,c}, \boldsymbol{\Sigma}_{r,c} : c = 1,2,...,C \}
    \label{eq:theta_parameters}
\end{equation}
where $\bd{M}_{s,c}, \boldsymbol{\Sigma}_{s,c}$ characterize the conditional distribution $p(\bd{s}_t|\bd{s}_{t-1})$ and $\bd{M}_{r,c}, \boldsymbol{\Sigma}_{r,c}$ the conditional distribution $p(\bd{r}_t|\bd{s}_{t})$. Such parameters can be obtained by generalizing the recursions in \eqref{eq:eta_i} and \eqref{eq:sigma_i} to multiple entities. Specifically, if $\bd{s}_{t_1}, \bd{s}_{t_2}, ..., \bd{s}_{t_n}$ are the vector of loads of $K$ entities obtained at times associated with calendar type $c \in \{1, 2, ..., C\}$, that is, \mbox{$c = c(t_1) = c(t_2) = ... = c(t_n)$}, and if $\bd{u}_{t_1}, \bd{u}_{t_2} , ..., \bd{u}_{t_n}$ are the corresponding feature vectors for parameters $\bd{M}_{s,c}$ , $\boldsymbol{\Sigma}_{s,c}$ as given in \eqref{eq:model1_MT} or for parameters \mbox{$\bd{M}_{r,c}$ , $\boldsymbol{\Sigma}_{r,c}$} as given in \eqref{eq:model2_MT}, then the estimators for the mean and covariance parameters are given by
\begin{align}
    \bd{M}_i = &  \bd{M}_{i-1} + \frac{(\bd{s}_{t_i} - \bd{M}_{i-1} \bd{u}_{t_i})  \bd{u}_{t_i}^\top \bd{P}_{i-1}}{\lambda + \bd{u}_{t_i}^\top \bd{P}_{i-1}\bd{u}_{t_i}} \label{eq:mean_MT}\\
    \bd{\Sigma}_i  = & \bd{\Sigma}_{i-1} \\
        \label{eq:sigma_MT}
    & - \frac{1}{\bd{\gamma}_i} \Big( \bd{\Sigma}_{i-1}  -  \frac{\lambda^2 ( \bd{s}_{t_i}- \bd{M}_{i-1}\bd{u}_{t_i})( \bd{s}_{t_i}- \bd{M}_{i-1}\bd{u}_{t_i})^\top }{(\lambda+\bd{u}_{t_i}^\top \bd{P}_{i-1}\bd{u}_{t_i})^2} \Big)  \nonumber 
\end{align}
with $\bd{P}_i$ and $\gamma_i$ given by \eqref{eq:P_i} and \eqref{eq:gamma_i}, respectively, $\bd{M}_0 = \bd{0}$, $\boldsymbol{\Sigma}_0 = \bd{0}$, $\bd{P}_0 = I_{K}$ and $\gamma_0 = 0$.


The equations above describe how to update the model parameters at each time $t$ and for each calendar type $c$. With these recursions, we obtain probabilistic models characterized by dynamic means and covariance matrices. In single-task learning, we update a vector $\boldsymbol{\eta}$ corresponding to a scalar mean consumption, while in multi-task learning we update the matrix $\bd{M}$ corresponding to a vector of mean consumptions. Such a matrix describes a linear combination of the $K$ entities at each time $t$. In addition, the covariance matrix $\boldsymbol{\Sigma}$ not only contains the variance as in single-task learning, but also provides the correlation between different tasks. Such parameters are updated adding corrections to the previous parameters $\bd{M}_{i-1}$ and $\boldsymbol{\Sigma}_{i-1}$. These corrections are proportional to the fitting error of the previous parameter $\bd{s}_{t_i} - \bd{M}_{i-1}\bd{u}_{t_i}$ so that parameters are updated according to their fit to the most recent data.

\subsection{Prediction}
The previous section details the recursive process of updating the HMM parameters using the most recent data. The following theorem demonstrates how to obtain probabilistic forecasts using these parameters.

\begin{theorem} 
\label{theorem}
Let $\left\{\bd{s}_t, \bd{r}_t \right\}_{t\geq 1}$ be an HMM characterized by parameters $\Theta$ defined in \eqref{eq:theta_parameters}. Then, for $i=1,2,...,L$
\begin{equation}
\label{eq:theorem}
    p(\bd{s}_{t+i}|\bd{s}_t, \bd{r}_{t+1:t+i}) = \set{N}(\bd{s}_{t+i}; \widehat{\bd{s}}_{t+i}, \widehat{\bd{E}}_{t+i})
\end{equation}
where mean $\widehat{\bd{s}}_{t+i}$ and covariance matrix $\widehat{\bd{E}}_{t+i}$ can be recursively obtained by
\begin{align}
    \widehat{\bd{s}}_{t+i} = & \bd{W}_1 (\bd{W}_1+ \bd{W}_2)^{-1}\bd{M}_{r,c}\bd{u}_r  + \nonumber \\
    &  \bd{W}_2 (\bd{W}_1+\bd{W}_2)^{-1} \bd{M}_{s,c}\widehat{\bd{u}}_{s} \label{eq:pred_mean}\\
   \widehat{\bd{E}}_{t+i} = & \bd{W}_2 (\bd{W}_1 +\bd{W}_2)^{-1}\bd{W}_1 \label{eq:pred_error}
\end{align}
$\text{with }\bd{W}_1 = \bd{\Sigma}_{s,c} + \bd{M}_{s,c}\bd{N} \widehat{\bd{E}}_{t+i-1}(\bd{M}_{s,c}\bd{N})^\top , \ \ \bd{W}_2 = \bd{\Sigma}_{r,c},$
\mbox{$\widehat{\bd{u}}_s = [1, \widehat{\bd{s}}_{t+i-1}]^\top, \bd{u}_r = u_r(\bd{r}_{t+i}), c  =  c(t+i),  
\bd{N} =
\begin{bmatrix}
0 \\
I_K
\end{bmatrix}.$}
In addition, the initial values are given by $\widehat{\bd{s}}_t = \bd{s}_t$ and $\widehat{\bd{E}}_t = \bd{0}$.  
\end{theorem}


The above theorem can be proven using induction together with the properties of HMMs. This theorem provides probabilistic load forecasts $\widehat{\bd{s}}_{t+i}$ and the estimates of their accuracy $\widehat{\bd{E}}_{t+i}$, for \mbox{$i=1,2,...,L$}. Every time new instance vectors $\bd{x}$ are available, these forecasts are obtained using the recursions \eqref{eq:pred_mean} and \eqref{eq:pred_error} with the most recent parameters. Specifically, the probabilistic forecast $\set{N}(\bd{s}_{t+i}; \widehat{\bd{s}}_{t+i}, \widehat{\bd{E}}_{t+i})$ at time $t+i$ for $i=1,2,...,L$ are obtained using the probabilistic forecast at previous time $\set{N}(\bd{s}_{t+i-1}; \widehat{\bd{s}}_{t+i-1}, \widehat{\bd{E}}_{t+i-1})$, observations vectors $\bd{r}_{t+i}$ at time $t+i$ and the model parameters corresponding to the calendar type at time $t+i$, denoted as $c=c(t+i)$.

The main distinction from single-task learning lies in how the probabilistic forecast for each entity is generated. In the proposed multi-task method, load forecasts of each entity are also based on models and predictions from other entities. Specifically, as shown in \eqref{eq:pred_mean}, the load forecast $\widehat{\bd{s}}_{t+i}$ is given by the linear combination of $\bd{M}_{r,c}\bd{u}_r$ and $\bd{M}_{s,c}\widehat{\bd{u}}_s$ that are given by models and previous predictions for all entities. In addition, the weights $\bd{W}_1$ and $\bd{W}_2$ are given by the relationships among entities represented by the covariance matrices $\boldsymbol{\Sigma}_{s,c}$ and $\boldsymbol{\Sigma}_{r,c}$. The covariance matrix $\widehat{\bd{E}}_{t+i}$ describes the estimated uncertainty in the forecasts and how forecasts for one entity are influenced by other entities. In particular, each diagonal term $(\widehat{\bd{E}}_{t+i})_{j,j}$ describes the variance for the forecasts for the $j$-th entity, while each non diagonal term $(\widehat{\bd{E}}_{t+i})_{j,k}$ describes the correlation between the forecasts for the $j$-th and $k$-th entities. 

The results in the previous sections describe the proposed method's learning and prediction steps. In the following section, we present the experimental results that validate the proposed techniques.

\section{Experiments}
\label{sec:experiments}

\begin{figure*}[h!]
    \centering
    \begin{subfigure}[b]{0.35\textwidth}
        \centering
        \psfrag{Load consumption [GW]}[b][][0.7]{Load consumption [GW]}
        \psfrag{Time [Hours]}[t][][0.7]{Time [Hours]}
        \psfrag{data1abcdefghijklmn}[l][l][0.6]{Load}
        \psfrag{data2}[l][l][0.6]{Proposed}
        \psfrag{data3}[l][l][0.6]{MTGP}
        \psfrag{data4}[l][l][0.6]{KMT}
        \psfrag{2.5}[][][0.6]{2.5}
        \psfrag{3}[][][0.6]{3}
        \psfrag{3.5}[][][0.6]{3.5}
        \psfrag{4}[][][0.6]{4}
        \psfrag{6}[][][0.5]{}
        \psfrag{12}[][][0.6]{12}
        \psfrag{18}[][][0.5]{}
        \psfrag{24}[][][0.6]{24}
        \psfrag{30}[][][0.5]{}
        \psfrag{36}[][][0.6]{36}
        \psfrag{42}[][][0.5]{}
        \psfrag{48}[][][0.6]{48}
        \includegraphics[width=\textwidth]{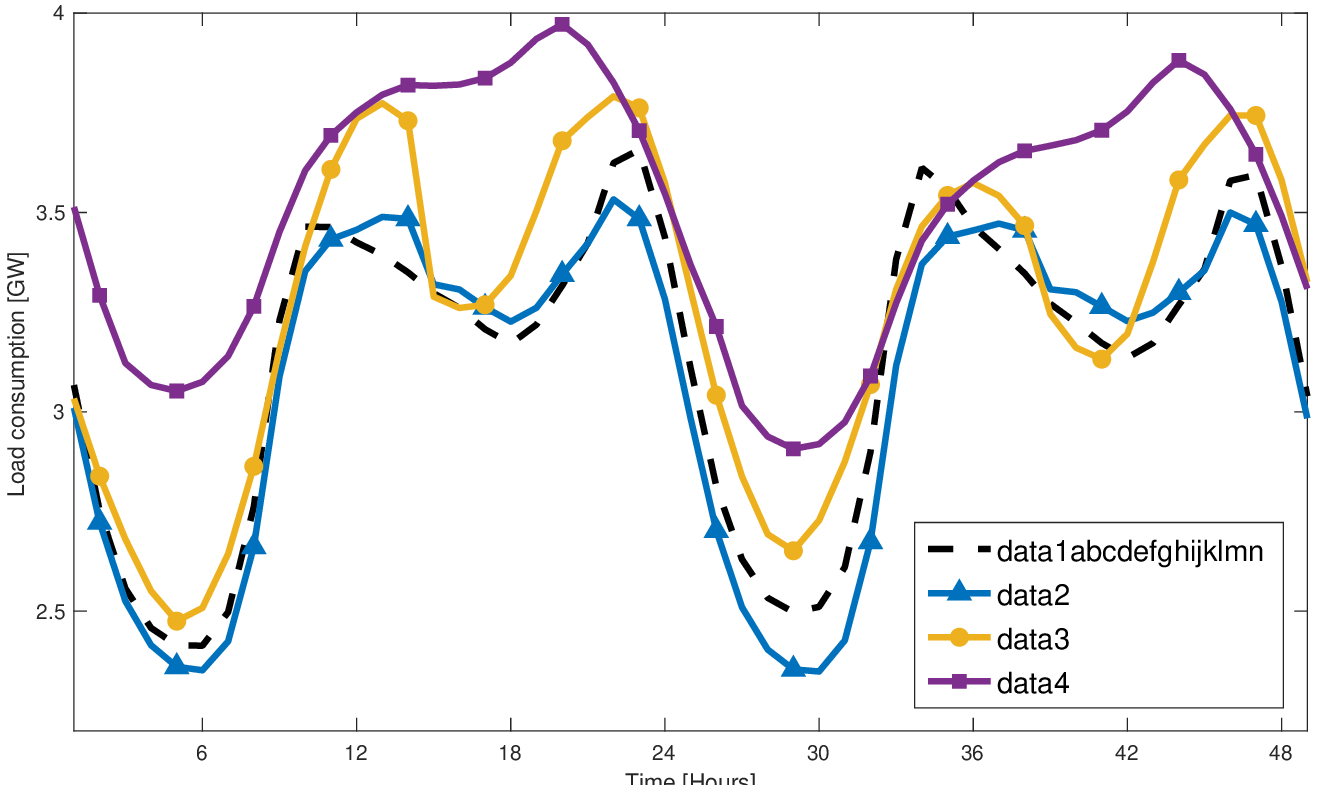}
        \caption{Load prediction for GEFCom2017 dataset.}
        \label{fig:gefcom2017_loads}
    \end{subfigure}
    \hspace{0.5cm}
        \begin{subfigure}[b]{0.35\textwidth}
        \centering
        \psfrag{CDF}[b][][0.7]{CDF}
        \psfrag{Error [GW]}[t][][0.7]{Error [GW]}
        \psfrag{data1abcdefghijklmn}[l][l][0.6]{Proposed}
        \psfrag{data2}[l][l][0.6]{MTGP}
        \psfrag{data3}[l][l][0.6]{KMT}
        \psfrag{1500}[][][0.6]{1.5}
        \psfrag{600}[][][0.6]{0.6}
        \psfrag{400}[][][0.6]{0.4}
        \psfrag{200}[][][0.6]{0.2}
        \psfrag{0.2}[][][0.6]{0.2}
        \psfrag{0.4}[][][0.6]{0.4}
        \psfrag{0.6}[][][0.6]{0.6}
        \psfrag{0.8}[][][0.6]{0.8}
        \psfrag{1}[][][0.6]{1}
        \psfrag{0}[][][0.5]{}
        \includegraphics[width=\textwidth]{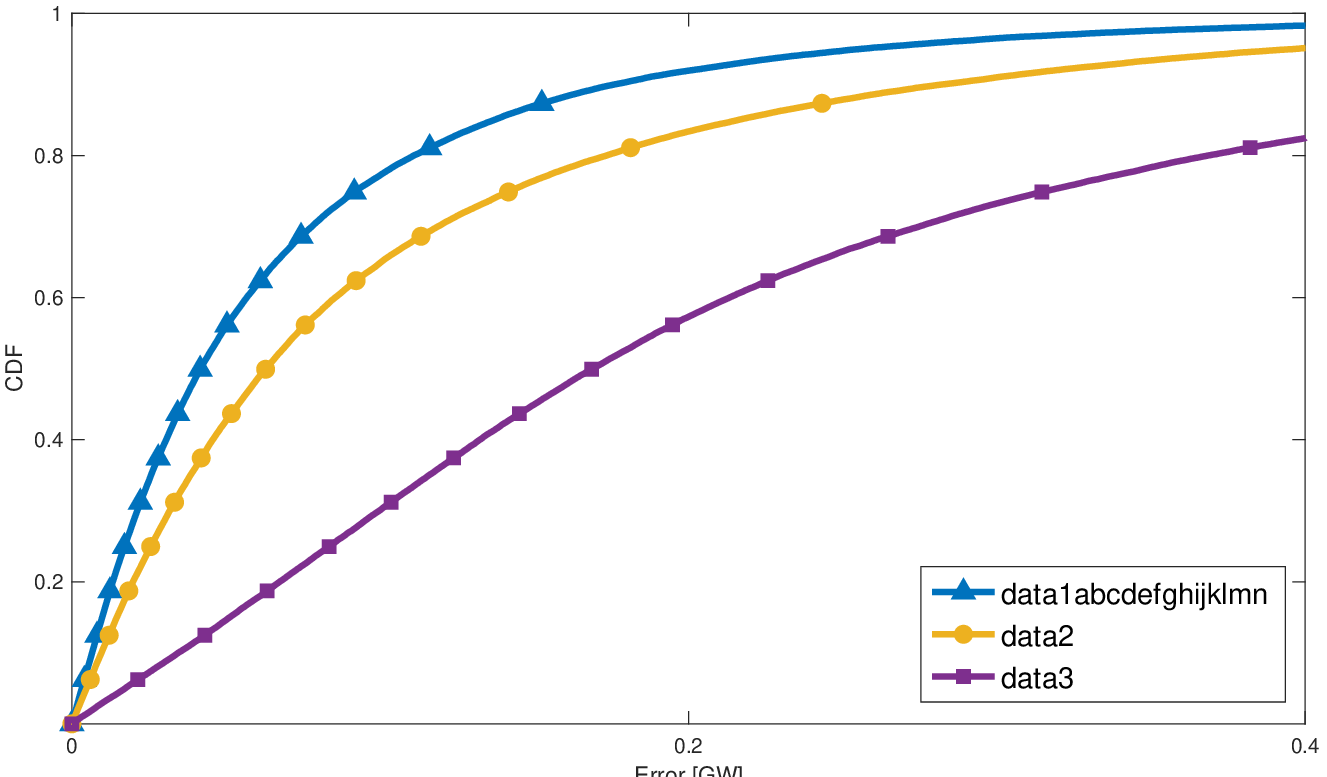}
        \caption{CDF for GEFCom2017 dataset.}
        \label{fig:gefcom2017_cdf}
    \end{subfigure}
    \caption{Load demand prediction and CDF in GEFCom2017 dataset.}
    \label{fig:pred_performance}
\end{figure*}

In this section, we first describe the datasets and the experimental setup used for the numerical results. Then, we compare the proposed method's performance with that of existing techniques.

\begin{table*}[]
\caption{MAPE and RMSE of prediction errors for the proposed method and state-of-the-art techniques.}
\tiny
\label{tab:experiments}
\resizebox{\textwidth}{!}{%
\begin{tabular}{cl|rrrr|rrrr|rr}
\toprule
 &          & \multicolumn{2}{|c}{\textbf{MTGP}}                            & \multicolumn{2}{c|}{\textbf{MTGP (online)}}                   & \multicolumn{2}{c}{\textbf{KMT}}                             & \multicolumn{2}{c}{\textbf{KMT (online)}}                    & \multicolumn{2}{|c}{\textbf{Proposed}}                        \\
&          & \multicolumn{1}{|c}{\textbf{MAPE}} & \multicolumn{1}{c}{\textbf{RMSE}} & \multicolumn{1}{c}{\textbf{MAPE}} & \multicolumn{1}{c|}{\textbf{RMSE}} & \multicolumn{1}{c}{\textbf{MAPE}} & \multicolumn{1}{c}{\textbf{RMSE}} & \multicolumn{1}{c}{\textbf{MAPE}} & \multicolumn{1}{c|}{\textbf{RMSE}} & \multicolumn{1}{c}{\textbf{MAPE}} & \multicolumn{1}{c}{\textbf{RMSE}} \\ \hline
\multirow{8}{*}{\rotatebox{90}{\textbf{GEFCom2017}}} & Entity 1 & 8.74                     & 0.35                     & 7.19                     & 0.30                     & 16.21                    & 0.65                     & 15.45                    & 0.64                     & \textbf{4.64}            & \textbf{0.22}            \\
& Entity 2 & 4.55                     & 0.07                     & 4.45                     & 0.07                     & 10.21                    & 0.14                     & 10.13                    & 0.15                     & \textbf{3.17}            & \textbf{0.05}            \\
& Entity 3 & 6.16                     & 0.24                     & 6.03                     & 0.24                     & 13.68                    & 0.45                     & 13.08                    & 0.46                     & \textbf{4.16}            & \textbf{0.17}            \\
& Entity 4 & 6.55                     & 0.11                     & 6.17                     & 0.10                     & 13.69                    & 0.20                     & 12.47                    & 0.20                     & \textbf{4.02}            & \textbf{0.07}            \\
 & Entity 5 & 6.05                     & 0.07                     & 5.86                     & 0.07                     & 13.79                    & 0.17                     & 13.77                    & 0.16                     & \textbf{4.08}            & \textbf{0.05}            \\
  & Entity 6 & 6.18                     & 0.14                     & 6.17                     & 0.14                     & 15.24                    & 0.32                     & 15.05                    & 0.33                     & \textbf{4.60}            & \textbf{0.11}            \\
 & Entity 7 & 5.38                     & 0.04                     & 5.42                     & 0.04                     & 17.92                    & 0.12                     & 14.71                    & 0.10                     & \textbf{4.54}            & \textbf{0.03}            \\
& Entity 8 & 6.15                     & 0.16                     & 6.09                     & 0.15                     & 13.85                    & 0.30                     & 13.14                    & 0.30                     & \textbf{4.21}            & \textbf{0.11}            \\
    & TOTAL    & 6.22                     & 0.15                     & 5.92                     & 0.14                     & 14.32                    & 0.29                     & 13.48                    & 0.29                     & \textbf{4.18}            & \textbf{0.10}            \\ \hline
\multirow{9}{*}{\rotatebox{90}{\textbf{NewEngland}}} & Entity 1 & 11.26                    & 0.14                     & 11.09                    & 0.15                     & 26.57                    & 0.29                     & 22.26                    & 0.26                     & \textbf{9.46}            & \textbf{0.12}            \\
 & Entity 2 & 6.71                     & 0.10                     & 6.66                     & 0.10                     & 20.16                    & 0.26                     & 16.24                    & 0.24                     & \textbf{5.27}            & \textbf{0.08}            \\
 & Entity 3 & 12.71                    & 0.07                     & 13.19                    & 0.08                     & 50.46                    & 0.22                     & 36.86                    & 0.18                     & \textbf{11.38}           & \textbf{0.06}            \\
 & Entity 4 & 6.79                     & 0.28                     & 6.57                     & 0.27                     & 21.29                    & 0.74                     & 18.44                    & 0.72                     & \textbf{5.47}            & \textbf{0.23}            \\
& Entity 5 & 8.67                     & 0.09                     & 8.72                     & 0.09                     & 20.62                    & 0.20                     & 19.60                    & 0.20                     & \textbf{6.97}            & \textbf{0.07}            \\
& Entity 6 & 7.29                     & 0.14                     & 7.33                     & 0.14                     & 24.46                    & 0.42                     & 21.51                    & 0.43                     & \textbf{6.01}            & \textbf{0.12}            \\
& Entity 7 & 9.78                     & 0.24                     & 11.13                    & 0.29                     & 20.74                    & 0.44                     & 18.64                    & 0.45                     & \textbf{7.59}            & \textbf{0.20}            \\
  & Entity 8 & 5.95                     & 0.21                     & 5.90                     & 0.20                     & 18.36                    & 0.52                     & 15.34                    & 0.50                     & \textbf{4.67}            & \textbf{0.15}            \\
 & TOTAL    & 8.14                     & 0.16                     & 8.83                     & 0.16                     & 25.33                    & 0.39                     & 21.11                    & 0.37                     & \textbf{7.10}            & \textbf{0.13}            \\ \hline
\multirow{6}{*}{\rotatebox{90}{\textbf{SMART}}}                     & Entity 1 & 40.49                    & 0.70                     & \textbf{38.37}                    & 0.75                     & 51.22                    & 1.02                     & 101.79                   & 0.99                     & 45.27                    & \textbf{0.66}            \\
 & Entity 2 & 87.31                    & \textbf{0.72}                     & 83.65                    & 0.73                     & 136.83                   & 0.79                     & 101.03                   & 0.87                     & \textbf{77.49}           & \textbf{0.72}            \\
& Entity 3 & 149.85                   & 0.53                     & \textbf{138.71}                   & \textbf{0.53}                     & 459.51                   & 0.80                     & 262.10                   & 0.64                     & 174.22                   & 0.54                     \\
 & Entity 4 & 87.03                    & 1.39                     & 83.16                    & 1.41                     & 79.10                    & 1.25                     & 86.47                    & 1.22                     & \textbf{59.98}           & \textbf{1.14}            \\
& Entity 5 & 52.63                    & \textbf{1.37}                     & \textbf{50.71}                    & 1.39                     & 339.01                   & 4.11                     & 131.90                   & 2.81                     & 59.26                    & 1.48                     \\
& TOTAL    & 83.46                    & 0.94                     & \textbf{78.92}                    & 0.96                     & 213.14                   & 1.59                     & 136.66                   & 1.30                     & 83.24                    & \textbf{0.91}      \\\bottomrule
\end{tabular}%
}
\end{table*}

\subsection{Datasets and experimental setup}
Three publicly available load consumption datasets are used in the numerical experiments: a dataset from New England made available by ISO New England organization that registers hourly load demand, electricity cost information, weather data and system load for the ISO New England Control Area and its 8 load zones; a dataset from the Global Energy Forecasting Competition 2017 (GEFCom2017) with load demand of 8 regions  and weather information; and a dataset of meter readings from 2016 of 5 homes in western Massachusetts (SMART), 
which collects electricity usage and generation data,  and weather information. 

The methods are trained using the initial 30 days of the dataset. Prediction for all the algorithms is done daily at 11 a.m., and the algorithms obtain future loads for a prediction horizon of $L=24$ hours. Therefore, the vectors of load forecasts contain predictions for the next $1$ to $L$ hours. The hyper-parameters for all methods can be selected using techniques like cross-validation across a grid of potential values. For simplicity, for the state-of-the-art methods we determine the hyper-parameters using the default values defined for each of them, and for the proposed method we inspect one dataset and then we apply the same values in all datasets.

The proposed method is implemented as follows. The instance vector $\bd{x}_t~=~[\bd{s}_t^\top, \bd{r}_{t+1}, \bd{r}_{t+2},...,\bd{r}_{t+L}]^\top$ is composed of past loads and observations. The observations vector $\bd{r}_t=[\bd{r}_{t,1}, \bd{r}_{t,2}, ..., \bd{r}_{t,K}]$ contains information about the temperature at time $t$ and the mean of past temperatures in each entity. For simplicity, the observations vector $\bd{r}_{t,k}$ for each entity $k~=~1,2,...,K$ is represented by the feature vector $u_r(\bd{r}_{t,k})$ as in \cite{alvarez2021probabilistic}, with $R = 3$.
The calendar information $c(t)$ ranges from 1 to 24 to denote the hours of weekdays, and from 25 to 48 to denote the hours of weekends and holidays, that is, $c(t) \in \{1, 2, ..., 48\}$. Therefore, the proposed method obtains parameters $\bd{M}_{s,c}, \boldsymbol{\Sigma}_{s,c}, \bd{M}_{r,c}, \boldsymbol{\Sigma}_{r,c}$ for $c \in \{1,2, ...,48\}$ calendar type. These parameters are updated using forgetting factors set to $\lambda_s = 0.8$ and $\lambda_r = 0.7$ in all calendar types and datasets.

The performance of the proposed method is compared with 2 state-of-the-art techniques: multi-task Gaussian Process \cite{Zhang2014} and multi-task kernel-based method \cite{fiot2016electricity}. As both techniques are offline learning approaches, we train the models using the first 25 days. An additional experiment is conducted for both methods to convert them into online learning techniques. For this experiment, we retrain the models every 6 months to try to adapt the methods to evolving consumption patterns.

The multi-task Gaussian Process (MTGP) method is implemented as described in \cite{Zhang2014}. Specifically, such a method is trained using load consumption from previous days. Instance vectors $\bd{x}$ are defined as a matrix of vectors of length $m$, that is, $\bd{x}_{t,k} = [s_{t-1,k},s_{t-2,k}, ..., s_{t-m,k}]$, for entities $k=1,2,...,K$. In particular, we use the loads from the previous $m=20$ hours to define the instance vectors.

The kernel-based multi-task learning method (KMT) is implemented as described in \cite{fiot2016electricity}. Specifically, this method is trained using seasonal effects that characterize electricity demand. The instance vector is given by $\bd{x}_{t}~=~[c_{t}, h_{t}, d_{t}]$, where $c_{t}$ is the day of the week (Monday to Sunday), $h_{t}$ is the hour of the day, and $d_{t}$ is the day of the year. In particular, we use the multiplicative input kernel to train the model.

\subsection{Numerical results}

The performance of the three load forecasting algorithms is evaluated using the root mean square error (RMSE) and the mean absolute percentage error (MAPE). Table \ref{tab:experiments} shows the RMSE and MAPE, assessing the overall prediction errors of the three methods and their adaptation to online learning. These results show that the adaptation to online learning methods mostly achieve higher accuracy than offline learning algorithms. KMT method makes the highest error in its predictions due to the fact that it only accounts for seasonal variables and ignores previous load consumption data or weather information. In addition, the retraining process of MTGP method slightly enhances its offline results. The proposed method achieves the best performance in almost all the entities of the datasets, and the lowest errors in the total prediction. 

The forecasting performance of the proposed method is shown in Figure~\ref{fig:pred_performance} for GEFCom2017 dataset. Figure~\ref{fig:gefcom2017_loads} illustrates the load demand and load forecasts for one entity over a two-day period. Figure~\ref{fig:gefcom2017_cdf} presents the empirical cumulative distribution functions (CDFs) of the absolute prediction errors. These results indicate that our method achieves higher accuracy compared to existing techniques. In particular, Figure~\ref{fig:gefcom2017_cdf} shows that the probability of high errors is low for the proposed method. For instance, the error for the proposed method is approximately 0.1 GW with a probability of 0.8, while the other methods show errors greater than 0.2 GW with the same probability.

The proposed method achieves more accurate results than existing multi-task load forecasting techniques. Such a method requires simple operations to update the parameters and obtains accurate load predictions. The results show that our method can adapt to dynamic changes in consumption patterns, even in datasets with significant variability in load demand.

\section{Conclusion}
\label{sec:conclusion}

The paper presents an online learning probabilistic method for multi-task load forecasting. In particular, the techniques presented extend existing methods developed for single-task scenarios. The proposed multi-task learning techniques can adapt to changes in consumption patterns, learn the relationship among multiple entities and asses load uncertainties. The parameters of the model are updated using a simple recursive algorithm, and the method provides probabilistic predictions with the most recent parameters. We detail the online learning and probabilistic prediction steps of our method for multiple entities. We compare the proposed method with existing state-of-the-art techniques for multi-task load forecasting and evaluate their prediction accuracy in datasets that register the load demand of multiple entities, and contain diverse dynamic consumption patterns. Experimental results show that the proposed method outperforms existing multi-task learning techniques in different load consumption scenarios.


\begin{thebibliography}{10}
\providecommand{\url}[1]{#1}
\csname url@samestyle\endcsname
\providecommand{\newblock}{\relax}
\providecommand{\bibinfo}[2]{#2}
\providecommand{\BIBentrySTDinterwordspacing}{\spaceskip=0pt\relax}
\providecommand{\BIBentryALTinterwordstretchfactor}{4}
\providecommand{\BIBentryALTinterwordspacing}{\spaceskip=\fontdimen2\font plus
\BIBentryALTinterwordstretchfactor\fontdimen3\font minus
  \fontdimen4\font\relax}
\providecommand{\BIBforeignlanguage}[2]{{%
\expandafter\ifx\csname l@#1\endcsname\relax
\typeout{** WARNING: IEEEtran.bst: No hyphenation pattern has been}%
\typeout{** loaded for the language `#1'. Using the pattern for}%
\typeout{** the default language instead.}%
\else
\language=\csname l@#1\endcsname
\fi
#2}}
\providecommand{\BIBdecl}{\relax}
\BIBdecl

\bibitem{wang2023probabilistic}
C.~Wang, Y.~Wang, Z.~Ding, and K.~Zhang, ``Probabilistic multi-energy load
  forecasting for integrated energy system based on bayesian transformer
  network,'' \emph{IEEE Transactions on Smart Grid}, 2023.

\bibitem{wang2023multitask}
J.~Wang, K.~Wang, Z.~Li, H.~Lu, H.~Jiang, and Q.~Xing, ``A multitask integrated
  deep-learning probabilistic prediction for load forecasting,'' \emph{IEEE
  Transactions on Power Systems}, vol.~39, no.~1, pp. 1240--1250, 2023.

\bibitem{wang2018review}
Y.~Wang, Q.~Chen, T.~Hong, and C.~Kang, ``Review of smart meter data analytics:
  Applications, methodologies, and challenges,'' \emph{IEEE Transactions on
  Smart Grid}, vol.~10, no.~3, pp. 3125--3148, 2018.

\bibitem{li2023residential}
Y.~Li, F.~Zhang, Y.~Liu, H.~Liao, H.-T. Zhang, and C.~Chung, ``Residential load
  forecasting: An online-offline deep kernel learning method,'' \emph{IEEE
  Transactions on Power Systems}, 2023.

\bibitem{kim2024spatial}
H.~J. Kim and M.~K. Kim, ``Spatial-temporal graph convolutional-based recurrent
  network for electric vehicle charging stations demand forecasting in energy
  market,'' \emph{IEEE Transactions on Smart Grid}, 2024.

\bibitem{zhang2021survey}
Y.~Zhang and Q.~Yang, ``A survey on multi-task learning,'' \emph{IEEE
  Transactions on Knowledge and Data Engineering}, vol.~34, no.~12, pp.
  5586--5609, 2021.

\bibitem{Zhang2014}
Y.~Zhang, G.~Luo, and F.~Pu, ``Power load forecasting based on multi-task
  gaussian process,'' \emph{IFAC Proceedings Volumes}, vol.~47, no.~3, pp.
  3651--3656, 2014.

\bibitem{gilanifar2019multitask}
M.~Gilanifar, H.~Wang, L.~M.~K. Sriram, E.~E. Ozguven, and R.~Arghandeh,
  ``Multitask bayesian spatiotemporal gaussian processes for short-term load
  forecasting,'' \emph{IEEE Transactions on Industrial Electronics}, vol.~67,
  no.~6, pp. 5132--5143, 2019.

\bibitem{fiot2016electricity}
J.-B. Fiot and F.~Dinuzzo, ``Electricity demand forecasting by multi-task
  learning,'' \emph{IEEE Transactions on Smart Grid}, vol.~9, no.~2, pp.
  544--551, 2016.

\bibitem{amjady2001short}
N.~Amjady, ``Short-term hourly load forecasting using time-series modeling with
  peak load estimation capability,'' \emph{IEEE Transactions on Power Systems},
  vol.~16, no.~3, pp. 498--505, 2001.

\bibitem{espinoza2007electric}
M.~Espinoza, J.~A. Suykens, R.~Belmans, and B.~De~Moor, ``Electric load
  forecasting,'' \emph{IEEE Control Systems Magazine}, vol.~27, no.~5, pp.
  43--57, 2007.

\bibitem{liu2015probabilistic}
B.~Liu, J.~Nowotarski, T.~Hong, and R.~Weron, ``Probabilistic load forecasting
  via quantile regression averaging on sister forecasts,'' \emph{IEEE
  Transactions on Smart Grid}, vol.~8, no.~2, pp. 730--737, 2015.

\bibitem{paarmann1995adaptive}
L.~D. Paarmann and M.~D. Najar, ``Adaptive online load forecasting via time
  series modeling,'' \emph{Electric Power Systems Research}, vol.~32, no.~3,
  pp. 219--225, 1995.

\bibitem{obst2021adaptive}
D.~Obst, J.~De~Vilmarest, and Y.~Goude, ``Adaptive methods for short-term
  electricity load forecasting during covid-19 lockdown in france,'' \emph{IEEE
  Transactions on Power Systems}, vol.~36, no.~5, pp. 4754--4763, 2021.

\bibitem{von2020online}
L.~Von~Krannichfeldt, Y.~Wang, and G.~Hug, ``Online ensemble learning for load
  forecasting,'' \emph{IEEE Transactions on Power Systems}, vol.~36, no.~1, pp.
  545--548, 2020.

\bibitem{laouafi2017online}
A.~Laouafi, M.~Mordjaoui, S.~Haddad, T.~E. Boukelia, and A.~Ganouche, ``Online
  electricity demand forecasting based on an effective forecast combination
  methodology,'' \emph{Electric Power Systems Research}, vol. 148, pp. 35--47,
  2017.

\bibitem{ba2012adaptive}
A.~Ba, M.~Sinn, Y.~Goude, and P.~Pompey, ``Adaptive learning of smoothing
  functions: Application to electricity load forecasting,'' \emph{Advances in
  Neural Information Processing Systems}, vol.~25, 2012.

\bibitem{yang2019bayesian}
Y.~Yang, W.~Li, T.~A. Gulliver, and S.~Li, ``Bayesian deep learning-based
  probabilistic load forecasting in smart grids,'' \emph{IEEE Transactions on
  Industrial Informatics}, vol.~16, no.~7, pp. 4703--4713, 2019.

\bibitem{alvarez2021probabilistic}
V.~{\'A}lvarez, S.~Mazuelas, and J.~A. Lozano, ``Probabilistic load forecasting
  based on adaptive online learning,'' \emph{IEEE Transactions on Power
  Systems}, vol.~36, no.~4, pp. 3668--3680, 2021.

\end{thebibliography}


\end{document}